\documentclass[twoside]{article}

\usepackage[accepted]{aistats2024}
\usepackage[round]{natbib}
\usepackage{hyperref}
\usepackage{bm}
\usepackage{graphicx}
\usepackage{enumitem}
\usepackage{amsmath, xparse}
\usepackage{amssymb}

\usepackage{algpseudocode}
\usepackage{ifthen}
\usepackage{algorithm}

\usepackage{stackengine}
\usepackage{xfp}
\usepackage{makecell}

\setstackgap{L}{.7\baselineskip}
\setstackgap{S}{1pt}
\def\prior#1#2{\ensurestackMath{#1}\stackanchor{pr}{#2}}
\def\post#1#2{\ensurestackMath{#1}\stackanchor{po}{#2}}
\newcommand{\mn}{\mathcal{M}}

\newcommand{\nth}[1]{#1^{\text{th}}}

\newcommand{\algorithmfootnote}[2][\footnotesize]{%
  \let\old@algocf@finish\@algocf@finish
  \def\@algocf@finish{\old@algocf@finish
    \leavevmode\rlap{\begin{minipage}{\linewidth}
    #1#2
    \end{minipage}}%
  }%
}

\usepackage{xcolor}

\makeatletter
\algnewcommand{\LineComment}[1]{\Statex \hskip\ALG@thistlm \(\triangleright\) #1}
\algnewcommand{\IndentLineComment}[1]{\Statex \hskip\ALG@tlm \(\triangleright\) #1}
\makeatother

\makeatletter

\newcommand{\splitatcommas}[1]{\begingroup\lccode`~=`, \lowercase{\endgroup
    \edef~{\mathchar\the\mathcode`, \penalty0 \noexpand\hspace{0pt plus 1em}}%
  }\mathcode`,="8000 #1%
  }

\begin{document}

%

%

\twocolumn[

\aistatstitle{Data Drift Monitoring for Log Anomaly Detection Pipelines}

\aistatsauthor{ Dipak Wani \And Samuel Ackerman \And Eitan Farchi \And Xiaotong Liu \And Hau-wen Chang \And Sarasi Lalithsena  }

\aistatsaddress{IBM Research}
]


\begin{abstract}
  Logs enable the monitoring of infrastructure status and the performance of associated applications. Logs are also invaluable for diagnosing the root causes of any problems that may arise. Log Anomaly Detection (LAD) pipelines automate the detection of anomalies in logs, providing assistance to site reliability engineers (SREs) in system diagnosis. Log patterns change over time, necessitating updates to the LAD model defining the `normal' log activity profile. In this paper, we introduce a Bayes Factor-based drift detection method that identifies when intervention, retraining, and updating of the LAD model are required with human involvement.  We illustrate our method using sequences of log activity, both from unaltered data, and simulated activity with controlled levels of anomaly contamination, based on real collected log data.
\end{abstract}

\section{Introduction
\label{sec:introduction}
}

Log anomaly detection in AIOps, short for Artificial Intelligence for Information Technology Operations, is a critical component of modern information technology (IT) operations. Log anomaly detection (LAD) pipelines are constructed by leveraging various machine learning techniques to analyze a large volume of logs and identify anomalous logs that may indicate potential issues or problems. However, applications and their infrastructure update over time, leading to changing log patterns. Thus, to accommodate these changing patterns, it is necessary to update the LAD pipeline accordingly to ensure its accuracy and effectiveness. Activity change detection is important because the users of the pipeline may not be aware of when to retrain it, and which data to include in the retraining.

Log activity change can generally be characterized as one of the following types:

\begin{itemize}
    \item \textbf{Anomaly}: a significant---but typically sporadic or not-sustained---deviation from the normal or expected distribution of activity.
    \item \textbf{Drift}: a systematic and sustained change in the distribution of log activity, possibly occurring gradually instead of suddenly.
\end{itemize}

 



Anomalies are sudden changes, often caused by errors, faults, rare occurrences, or temporary unusual behavior of the application.  In contrast, drift can occur due to various factors such as updating the application version, adding new features, infrastructure changes, removing old features from the application, and so on. A key difference is that in sustained drift, the changed observed logs reflect a new `normal' activity profile, and thus should be used to update or retrain the LAD model to avoid (false) detection of newly changed (but normal) activity as anomalies.  However, the distinction between drift and anomalies may be subjective, and we often use the term `drift detection' to denote change detection in general.   



In this paper, we propose a pipeline to detect data drift, recommend when to retrain the model, and differentiate between anomalies and data drift. Our approach is based on modeling the expected frequencies of various log patterns (`count vectors') and determining the Bayes Factor (BF) to statistically measure observed deviation from the expected baseline of the LAD pipeline.  If the distribution shifts from the baseline learned during training, it can raise an alert for drift detection. We ask for an intervention by the user to determine whether to retrain the model.


Section \ref{sec:related_work} discusses the related work on LAD pipelines, drift detection models based on Bayesian inference, and other approaches. In Sections \ref{sec:problem_setup} and \ref{sec:bayes_factor} we describe the mathematical setup, define the problem, and the drift detection algorithm. We discuss the experiments performed and the results achieved in Section \ref{sec:experiments}.

\section{Related Work}
\label{sec:related_work}
With respect to machine learning (ML) models, drift can be defined as a change between the real-time production data and the data used to train the model initially. Most well-known techniques \cite{gemaque2020overview} for detecting drifts involve comparing the different distributions. Concept drift detection \citep{gama2014survey} focuses on identifying changes between the observed and learned decision boundary, which indicates that the learned relationship between the model inputs and the target feature(s) has shifted. Prediction drift focuses on the model's output or the label distribution. Meanwhile, data (or feature) drift \cite{lu2018learning} focuses on the model's input data. Drifts of the predicted value are a good indicator of detecting the drift. However, in an actual production environment ground truth data about the predicted values will not always be available in real time. Hence, we focus on the data drifts in our problem setting. 

Existing work on drift detection considers statistical methods and ML-based approaches. For instance, \cite{lewis2022augur} and \cite{zhou2019framework} leverage the Kullback-Leibler divergence; \cite{lewis2022augur} and \cite{cieslak2009framework}  leverage Kolmogorov-Smirnov Statistic to detect drifts. \cite{Ackerman:2021} uses the BF to detect drifts of API usages, modeled as a categorical distribution, which can help in determining if the usage patterns of the system have shifted. \cite{zenisek2019machine} presents an example of an ML-based approach to detect drifts; such approaches typically use an ML model to either classify the system state as normal or not (an increase in the instances classified as `abnormal' may indicate drift), or to predict future values of the monitored data stream (increased divergence between forecasts and observed values may indicate the original model has drifted). While leveraging the benefits of both statistical and ML-based approaches, \cite{baburouglu2021novel} proposes a hybrid approach to drift detection. 

Log data captures the system state and its run-time behaviors. Logs are often used as a source to detect system failures and anomalies. Dynamic changes in the states of various individual system services may be reflected in changes in their respective logs. These dynamically changing logs will have an impact on the accuracy of the offline trained models unless they are retrained to capture the new logs. The proposed approach for drift detection of logs is inspired by \cite{Ackerman:2021} and applies the BF to detect the drifts in log data, specifically on the extracted summary feature vectors.
\section{Dataset
\label{sec:dataset}}


\subsection{Data collection
\label{ssec:data_collection}
}

An LAD model or pipeline is tailored for a specific system application. The model estimates a profile of the baseline `normal' log activity to be estimated, which requires initial data collection; new activity, which may be different in nature from this baseline, will be compared to the baseline.

The data we used for our experiments was collected by aggregating logs from different micro-services within a certain deployment.  Due to limited access to log data across applications and time, we could access log data for a deployment that was captured in 2020 and 2022. In our experiments, we considered ~300k logs each from 2020 to be the normal ($N$) baseline, and from 2022 to be the potential anomalies ($A$).  We found 5 microservices for which data were available for both years: MTR, NLU, MMR, STR, and STRS (actual names masked for confidentiality).

\subsection{Data feature extraction
\label{ssec:data_feature_extraction}}
System logs are heavily textual data type; they usually display information like API calls, HTTP error messages, etc.  Because of this, the logs need to be preprocessed---in our case, converted to a numeric vector form---before feeding them to any anomaly detection model.  We use a LAD pipeline described in \cite{An:2022} and \cite{Xu-count-vectors:2009} that uses a principal components analysis (PCA)-based method to detect anomalies in the logs. The pipeline is summarized by the following phases:
\begin{enumerate}
\item Template generation
\begin{enumerate}
    \item Preprocess the training logs by cleaning them and converting them to a standard format. Cleaning steps include removing empty messages, masking timestamp values, and removing common prefix patterns.
    \item Parse these logs using the tree-based template algorithm DRAIN \cite{he-drain:2017} which generates \textbf{templates} by extracting the most frequent logs. Each template is a regular expression pattern. For example, \textit{$<$*$>$ Error obtaining quote: $<$*$>$ [object Object]} is a template where \textit{$<$*$>$} is a pattern which matches any word.
\end{enumerate}
    \item Feature vector generation
    \begin{enumerate}
     \item Next, for the same dataset of training logs, we sort the logs by time and generate a chunk of logs for every 10-second window. We call these chunks `windowed logs'. 
    \item We match each line in the windowed logs to a template using regex pattern matching.  A particular log line not matching any template is assigned to one of the following `unknown' template labels: $unk_{error}$ if the log line contains any error keywords, and  $unk_{normal}$ otherwise.  A feature vector is extracted for each chunk of windowed logs; the features correspond to each of the identified templates, with the value being the count of log line matches in that window to the template.  This is called the `count vector'.   
    \end{enumerate}

\end{enumerate}

The feature extraction procedure captures the log activity in terms of the frequency of various textual templates while discarding finer-resolution textual information. For example, if a template contains the $<$*$>$ pattern, the actual values that match this pattern are ignored.  This method is discussed in detail in section 4.2 of \cite{liu:2020}. Note also that the template extraction (i.e., data collection) is conducted separately for each application, which may have uniquely formatted log printouts, and thus a unique set of templates and activity profiles.

\section{Problem Setup } \label{sec:problem_setup}
In this section, we describe the methods used to model the logs as feature vectors and detect drift.

\subsection{General Notation
\label{ssec:notation}
}

We first define the notation used in the remainder of the paper.  Vector-valued objects are denoted in bold, like $\boldsymbol{X}$, with length $|\boldsymbol{X}|$.   $E(\boldsymbol{X})$ represents the result of dividing a nonnegative-valued vector $\boldsymbol{X}$ elementwise by the sum of element values (i.e., converting its elements to probabilities).   $X=(a,b,c)$ denotes an ordered sequence of elements $a,b,c$.  Both vectors $\boldsymbol{X}$ and sequences $X$ are 1-indexed.  Depending on the context, $X[i]$ and $\boldsymbol{X}[i]$ indicate the $\nth{i}$ element of the list or vector; similarly, $X[i:]$ and $\boldsymbol{X}[i:]$ mean the list or vector beginning with the $\nth{i}$ element, dropping the first $i-1$.  $\sum\boldsymbol{X}$ is shorthand for the scalar sum of the elements of $\boldsymbol{X}$ (i.e., $\sum_{i=1}^{|\boldsymbol{X}|}\boldsymbol{X}[i]$).


\subsection{Count vector feature construction
\label{ssec:count_vector_construction}
}

As noted in Section~\ref{ssec:data_feature_extraction}, the first data processing step is the identification of a set of unique templates (of size $K\geq 1$) in the baseline training sample of logs.  An observed log line in testing samples may match to one of these $K$, or one of the two `unknown' labels.  In our data, the number of identified templates were 99 (NLU), 83 (MTR), 42 (STRS), 62 (MMR), and 66 (STR).

Let $\boldsymbol{C}$ denote an arbitrary count vector, and $\boldsymbol{C}_t$ denote that corresponding to the $\nth{t}$ 10-second-long window of logs.  As each $\boldsymbol{C}$ measures the frequencies of template matches in a given window, they are each a nonnegative integer-valued vector of length $K+2$; each index $i,\:i=1,\dots,K$ (i.e., $\boldsymbol{C}[i]$) corresponds to the $\nth{i}$ template identified in the training data, while the final indices $K+1,K+2$ count matches to the two unknown types.  The sum of a count vector's elements, $\sum\boldsymbol{C}$, thus equals the number of log lines observed in a window.  Because the windows are fixed 10-second time spans, the number of logs in a given window, and thus the vector sum, can vary.

For a particular application, let $C_N=(\boldsymbol{C}^N_1,\dots,\boldsymbol{C}^N_n)$ and $C_A=(\boldsymbol{C}^A_1,\dots,\boldsymbol{C}^A_m)$ be collected count vectors under the 2020 non-anomalous ($N$) and 2022 anomalous ($A$) settings.  By definition, each training sample vector $\boldsymbol{C}\in C_N$ must have $\boldsymbol{C}[K+1]=\boldsymbol{C}[K+2]=0$, since each training sample log must by definition match to a template.  For each test $\boldsymbol{C}\in C_A$, there are no such restrictions on the values; however, in our collected data (Section~\ref{ssec:data_collection}) we observed for many of these vectors $\boldsymbol{C}\in C_A$ that there was no overlap with the identified templates, and that all were of the unknown type.

In our experiment (Section~\ref{ssec:cv_drift_simulation}), we simulate distribution change from normal to anomalous by generating count vectors that are mixtures of the two samples $C_N$ and $C_A$.  If many count vectors in $C_A$ do not have templates that appeared in $C_N$, this means the two distributions do not have significant overlap, meaning change detection should be easier for our algorithm.  






\subsection{Multinomial distribution
\label{ssec:multinomial}
}

As count vectors $\boldsymbol{C}$ measure the observed frequency of templates (or unknowns) in a time window, it is sensible to consider the distribution of these templates.  Let $\boldsymbol{P}=\begin{bmatrix}p_1 & \dots &p_{K+2}\end{bmatrix}$ and $p_i\in[0,1],\:\forall i=1,\dots,|\boldsymbol{P}|$ and $(\sum\boldsymbol{P})=1$ be a $(K+2)$-length vector of probabilities.  A multinomially ($\mn$)-distributed vector $\boldsymbol{X}$, $\boldsymbol{X}\sim \mn(\boldsymbol{P}, n)$ denotes the frequencies of $n$ items (i.e., log lines) that belong to one of $K+2$ categories (i.e., templates), each with probability $\boldsymbol{P}[i]$; each of the $n$ items' assignments is independent and identically-distributed (the $\boldsymbol{P}[i]$ are fixed).
Let $\mn(\boldsymbol{P},\cdot)$ denote multinomial samples where the item total $n$ (i.e., the number of log lines) is not fixed a priori, which is appropriate for our scenario (see Section~\ref{ssec:count_vector_construction}).  

Our LAD model is based on the assumption that the template frequencies sufficiently capture some sort of stable characteristic behavior of the baseline sample.  Thus, we model each $\boldsymbol{C}\sim \mn(\boldsymbol{P},\cdot)$.
This assumption would be violated if the log lines are correlated (not independent) in some way or if the probabilities are not constant over the time window. 

Given a set of count vectors $C$ of equal length $m$, but each having possibly different sum $n_j,\:j=1,\dots,|C|$, the maximum likelihood estimate (MLE) of $p_i$ is given by Equation~\ref{eq:multinomial_mle}:
\begin{equation}
\label{eq:multinomial_mle}
\hat{p}_i=\frac{\sum_{j=1}^{|C|}C[j][i]}{\sum_{k=1}^{|C|}n_k},\:i=1,\dots,m
\end{equation}

We can therefore apply the chi-squared test (\citealt{multinomial_tests}, Equation 6; \citealt{bain1992intro}) shown in Equation~\ref{eq:pvalue} to test whether the count vectors in $C$ seem to follow a common multinomial distribution with MLE probability vector $\boldsymbol{P}=\begin{bmatrix}\hat{p}_1&\dots&\hat{p}_m\end{bmatrix}$.

\begin{equation}
\label{eq:pvalue}
X=\sum_{j=1}^{|C|}\sum_{i=1}^m\frac{(C[j][i] - \hat{p}_i n_j)^2}{\hat{p}_i n_j} \quad\sim\chi^2_{(m-1)(|C|-1)}
\end{equation}

We perform this test separately for each system's year 2020 and 2022 count vector samples $C_N$, $C_A$, where $\boldsymbol{P}_N$ and $\boldsymbol{P}_A$ are the respective MLE vectors.  For all cases, except for STR 2020, the p-value is nearly 1.0 (highly insignificant), indicating the samples seem to have a shared multinomial parameter. 
Thus it is reasonable to model $\boldsymbol{C}\sim\mn(\boldsymbol{P}_N,\cdot)$ or $\mn(\boldsymbol{P}_A,\cdot)$ for $\boldsymbol{C}$ in $C_N$ or $C_A$, respectively.

\subsection{Count vector drift simulation
\label{ssec:cv_drift_simulation}}
As mentioned in section \ref{ssec:count_vector_construction}. count vectors from 2022, $\boldsymbol{C}\in C_A$ had $\leq1\%$ overlap with templates from $C_N$. We assume, that this must have been because of the gradual change in application logs over 2 years. To simulate the different stages of drift that may have occurred from 2020 to 2022, we perform simulations in the count vectors. 
Since it is difficult to explicitly model the different types of drifts that may occur, we simulate them by injecting controllable amounts of anomalous observations, as follows.



Drift can be characterized as a significant change in the observed (estimated posterior) $\boldsymbol{P}$ relative to the expected normal baseline $\boldsymbol{P}_N$, and this statistical assumption underlies our detection procedure (Section~\ref{ssec:windowed_BF}).  In simulating drift, we do not explicitly generate multinomial samples according to a given $\boldsymbol{P}\ne\boldsymbol{P}_N$ (the statistical assumption), but rather use the observed vectors in $C_A, C_N$ in a controllable manner, as follows.  Let $\textrm{draw}(X)$ denote an item drawn at random with uniform probability from the set of items $X$.  The simulation procedure is shown in Algorithm~\ref{alg:count_vector_sample}, in which the returned count vector $\boldsymbol{C}$ with contamination $p\in[0,1]$ represents a window where $100p\%$ of log lines (not time) are anomalous.

\begin{algorithm}
    \caption{Drifted count vector simulation\label{alg:count_vector_sample}}
    \begin{algorithmic}[1]
    \Procedure{SimDrift}{
        $C_N,C_A,p$}
        \IndentLineComment $p\in[0,1]$ is the proportion of anomalous ($A$) lines to inject
        \State $\boldsymbol{C}_N\gets\textrm{draw}(C_N)$
        \State $\boldsymbol{C}_A\gets\textrm{draw}(C_A)$
        \State $n_N \gets \sum \boldsymbol{C}_N$
        \State $n_A \gets \sum \boldsymbol{C}_A$
        \State
        $\boldsymbol{C}\gets (p\times E(\boldsymbol{C}_A) + (1-p)\times E(\boldsymbol{C}_N))$
        \State \Return $\boldsymbol{C}$
    \EndProcedure
\end{algorithmic}
\end{algorithm}

\subsection{Simulation settings
\label{ssec:simulation_settings}
}

In a single simulation,
$T=1000$ count vectors $C=(\boldsymbol{C}_1,\dots,\boldsymbol{C}_T)$ are drawn, each by an independent call of SimDrift (Algorithm~\ref{alg:count_vector_sample}) with the first two inputs fixed.  The simulation is repeated $R=50$ times at each combination of parameters in  Table~\ref{tab:simulation_settings}, consisting of a contamination level $p$ and length $\ell>1$. Contamination starts at window $t_s=501$.
Thus, in each sequence, the first $t_s-1=500$ are non-contaminated, drawn by setting $p=0.0$.  The final $T-t_s+1=500$ are drawn with time-varying contamination levels $p_t,\:t=t_s,\dots,T$, with contamination at level $p>0$ for windows $t=t_s,\dots,(t_s+\ell-1)$; `full' contamination means the contamination lasts until $T$, while `short' means the contamination returns to 0 after $\ell$ windows.  Roughly speaking, the full contamination is a sustained change, meant to mimic the `drift' change, while short, contamination which disappears is mean to mimic the `anomaly' change (Section~\ref{sec:introduction}).  

\begin{table}
\begin{tabular}{|p{8em}|p{15em}|}%
\hline
 Contamination $p$ & Contamination length $\ell$\\ 
 \hline
 \begin{itemize}[noitemsep,topsep=0pt]
    \item low: $p=0.1$
    \item high: $p=0.3$
\end{itemize} &
\begin{itemize}[noitemsep,topsep=0pt] \item short: $\ell=300$
 \item full: $\ell=\infty$
 \end{itemize}
 \vspace{2mm}
 
 $p_t=\begin{cases} 0.0, & t=1,\dots,t_s-1\\
 p, & t=t_s,\dots,t_s+\ell-1\\
 0.0, & t\geq t_s+\ell
 \end{cases}$\\
 \hline
\end{tabular}
\caption{Simulation settings.  $R$ repetitions are done at each combination of the input setting values.
\label{tab:simulation_settings}}
\end{table}

 


The simulation rationale is thus: 
A small $p>0$ represents a small change in the simulated template probabilities $\boldsymbol{P}$ relative to the baseline $\boldsymbol{P}_N$; in our case, this is particularly an introduction of \textit{previously unseen} templates.
A detection procedure is likely to be less sensitive when $p$ is smaller, and thus is less likely to detect this drift. Furthermore, a shorter contamination length $\ell$ is likely to have a lower cumulative effect on the detector.

\begin{itemize}[noitemsep]
    \item Reduce the likelihood of \textit{false positives}.
    \item Reduce the likelihood of \textit{false negatives}, that is not detecting contamination after a significant amount of time since $t_s$ when there was significant contamination.
    \item Given that $t_*\geq t_s$ (true positive), the detection delay $\delta=t_*-t_s$ should be low.
\end{itemize}

In our simulations, a large number of non-drifted and drifted samples are used to sufficiently measure both the false positive and negative rates, to give it a chance to `fail' in either scenario.  If drift is detected, we may want to take corrective action, such as re-training the LAD pipeline, which has a real cost.  A low false positive and low false negative rate and low detection delay together ensure that corrective action is reliably and quickly taken to mitigate any harm due to presence of drift.  See \cite{Ackerman:2021} and \cite{ackerman2020detection} for similar discussion.

\section{Bayes Factor
\label{sec:bayes_factor}}

\subsection{Background
\label{ssec:bf_background}}

In our setting, we model the multinomial probability vector $\boldsymbol{P}$ of the various log templates occurring.  The initial (or \textit{prior}) estimate, is based on the baseline non-anomalous sample estimate $\boldsymbol{P}_N$, reflecting the fact that, absent new information, we expect new observations to follow this same distribution.  The \textit{posterior} estimate is the prior estimate updated with new observed count vectors as they arrive.  We wish to compare two hypotheses: 1) the null ($H_0$) that the observed is similar to the prior (baseline), and 2) the alternative ($H_A$) that the posterior is significantly different, that is that there is drift from the prior.


The Bayes Factor (BF) is a statistical measure that can be used to compare the relative evidence in favor 
of these two hypotheses.  The BF at time $t$, denoted $BF_t$, is calculated as the ratio of the posterior and prior odds, that is, the relative statistical (multinomial distribution) fit of the data, using the posterior and prior estimates of the parameter of interest, here $\boldsymbol{P}$.



Thus, if the BF is close to 1, it means the posterior distribution looks similar to prior and thus, we conclude that the data is not drifted.  Higher values of the BF correspond to more evidence against the prior, that is, that there is drift in the log distribution.  The criterion for detecting drift is if $BF_t >1/\alpha$ (or $\ln{(1/\alpha)}$ for the log-BF), where $\alpha\in(0.0, 1.0]$ is a pre-determined level of statistical confidence against false positives.  

\subsection{Construction of windowed Bayes Factors
\label{ssec:windowed_BF}}

As noted in Section~\ref{ssec:cv_drift_simulation}, in the drift simulations each count vector $\boldsymbol{C}_t$ is created by constructing a $p_t$-weighted average of normalized count vectors drawn from the samples $C_A,C_N$, rather the \textit{synthetically drawing a multinomial sample} using the weighted average vector $p_t\boldsymbol{P}_A + (1-p_t)\boldsymbol{P}_N$.  However, our detection procedure will model $\boldsymbol{C}_t$s as having a multinomial distribution; this is how it will work on deployed data (Section~\ref{ssec:nonsimulated}
) where count vectors are observed in the field and not simulated.  

The detection algorithm assumes the observed sequence $C$ follows a fixed multinomial distribution with some parameter $\boldsymbol{P}$ (which a posterior $\post{\boldsymbol{P}}{}$ estimates) and compares it to the baseline $\boldsymbol{P}_N$ (represented by a prior $\prior{\boldsymbol{P}}{}$).
The prior assumption\footnote{Actually, a small adjustment to the MLE in Equation~\ref{eq:multinomial_mle} is made so that
after calculating the MLE $\boldsymbol{P}_N$, any non-observed templates indexed $i$, for which $\boldsymbol{P}_N[i]=0$ receive a small nonzero prior weight $\epsilon$ so that the posterior ratio is they are observed is not infinite.  The Dirichlet prior vector $\prior{\boldsymbol{P}}{}$ used for $\boldsymbol{P}_N$ is equal to $\kappa \boldsymbol{P}_N$ (after adjustment) where $\kappa>0$ controls the weight given to the prior evidence.} is that $\boldsymbol{C}_t\sim \mn(\boldsymbol{P}_N,\cdot),\:\forall t$.  The BF (Section~\ref{ssec:bf_background}) reflects the comparison of the current posterior estimate to the initial hypothesis of $\boldsymbol{P}_N$.

Given a time-ordered sequence of count vectors $C=(\boldsymbol{C}_1,\boldsymbol{C}_2,\dots)$, we apply a sequential monitoring procedure to calculate a sequence of BFs $(BF_1,BF_2,\dots)$.  In a real production pipeline, the count vectors will arrive in an online stream rather than being given at the beginning.
Note that normalization (line~\ref{line:count_vector_normalization}) to sum 1 is performed to give each window equal weight in the BF calculation rather than have it
be affected by the number of log lines in the window; simulated drifted vectors from Section~\ref{ssec:cv_drift_simulation} are already normalized.
Ordinarily, $BF_t$ reflects the information in the full history $(\boldsymbol{C}_1,\dots,\boldsymbol{C}_t)$.  We implement windowing (Algorithm~\ref{alg:windowed_bf}) parameterized by an integer window size $w\geq 1$, whereby $BF_t$ will reflect only the subset $(\boldsymbol{C}_{\textrm{max}(1,t-w+1)},\dots,\boldsymbol{C}_t)$ of at most the past $w$ observations; we use $w=100$ in our simulations in Section~\ref{ssec:simulation_settings}.  Windowing allows the BF to forget some previous observed $\boldsymbol{C}$s, thus reducing the influence of incidental but not sustained anomalous observations on future BFs.  Also, we set a grace period $g=100$ so that any significant BFs $BF_t,\:t< g$ are ignored, due to initial instability in the BFs due to the small window sizes.

Each $BF_t$ is the ratio of likelihoods of the posterior and prior estimates of the multinomial distribution $\boldsymbol{P}$.  We use a recursive formula from \cite{lindon2022multinomial} (Equation 27) in Algorithm~\ref{alg:multinomial_loglike}, which allows efficient calculation of the multinomial likelihoods without the gamma function inputs overfilling.  

\begin{algorithm}
    \caption{Gamma constant in multinomial log-likelihood (see \cite{lindon2022multinomial}, Equation 27)\label{alg:multinomial_loglike}}
    \begin{algorithmic}[1]
    \Procedure{LG}{
        $\boldsymbol{X}$}
        \State \Return $\displaystyle{\sum_{i=1}^{|\boldsymbol{X}|}\ln{\left(\Gamma(\boldsymbol{X}[i])\right)} - \ln{\left(\sum_{i=1}^{|\boldsymbol{X}|}\Gamma(\boldsymbol{X}[i])\right)}}$
    \EndProcedure
    \end{algorithmic}
\end{algorithm}

\begin{algorithm}[h!]
    \caption{Windowed BF algorithm \label{alg:windowed_bf}}
    \begin{algorithmic}[1]
    \Procedure{WindowedBF}{$C=(\boldsymbol{C}_1,\boldsymbol{C}_2,\dots)$}
        \LineComment{$\prior{\boldsymbol{\alpha}}{}$: Prior multinomial vector of non-drifted logs}
        \LineComment{$\boldsymbol{C}$: sequence of observed count vectors}
        \LineComment{$w$: integer, $w\geq 1$ window size to use in drift detection}
        \LineComment{$B_0$: log prior odds on log-BF}
        \State $W\gets (\boldsymbol{0})$ \Comment{Window of count vectors to keep}
        \State $W_S\gets (\prior{\boldsymbol{\alpha}}{})$ \Comment{Cumulative sums of $W$}
        \State $B\gets ()$ \Comment{log-BF sequence}
        \State $\boldsymbol{\theta}_0 \gets E(\prior{\boldsymbol{\alpha}}{})$
        \For{$t=1\textrm{ \textbf{to} }|C|$}
            \State $\tilde{\boldsymbol{C}}_t\gets E(\boldsymbol{C}_t)$ \label{line:count_vector_normalization}
        \If{$t>w$}
            \IndentLineComment{Drop element outside of window $w$}
            \State $W \gets W[\:2:\:]$
        \EndIf
        \For{$i=2\textrm{ \textbf{to} }|W_S|$} 
            \IndentLineComment{Windowed cumulative estimate of $\post{\boldsymbol{\alpha}}{i}$}
            \State $W_S[i] \gets W_S[i-1] + W[i-1]$
        \EndFor
        \LineComment{Use Algorithm~\ref{alg:multinomial_loglike} for log-BF at $t$ in $W$}
        \State \begin{align*}
        B[t]\gets B_0+\sum_{i=1}^{|W|}\biggl(&LG({W}_S[i] + W[i]) - LG(W_S[i])\\
        -&\sum_{j=1}^{|\boldsymbol{C}_t|}\ln{(W[i][j])}\boldsymbol{\theta_0}[j]\biggr)
        \end{align*}
        \LineComment{Append newest count vector to window}
        \State $W[\:|W|+1\:]\gets \tilde{\boldsymbol{C}}_t$
    \EndFor
    \State \Return $B$
    \EndProcedure
    \end{algorithmic}
\end{algorithm}

\section{Experiments
\label{sec:experiments}}

\subsection{Setup and evaluation
\label{ssec:setup_and_evaluation}
}

Using collected count vectors $C_N$ (from 2020) and $C_A$ (from 2022) from several different microservices (Section~\ref{ssec:data_collection}), we simulate drift as shown in Section~\ref{ssec:simulation_settings}; two values of two parameters $p$ and $\ell$ gives 4 total scenarios.  Setting $\alpha=0.05$ as the statistical confidence, the detection threshold is $c=\ln{(1/\alpha)}\approx 3$ for the log-BF.  For the $\nth{r}$ simulation run out of $R$, let $d_r\in\{0\}\bigcup\{g,\dots,T\}$ be the detection window, i.e., the first $t\geq g$ after the grace period such that $\ln{(BF_t)} > c$), or 0 if no detection is made by time $T$.  A correct decision occurs if $d_r \geq t_s$.  Following Section~\ref{ssec:simulation_settings}, define the following averages estimated across simulations $r$:

\begin{itemize}
    \item False positive rate (FPR): $\frac{1}{R}\sum_{r=1}^RI(g\leq d_r < t_s)$
    \item True positive rate (TPR): $\frac{1}{R}\sum_{r=1}^RI(d_r \geq t_s)$
    \item False negative rate (FNR): $\frac{1}{R}\sum_{r=1}^RI(d_r = 0)$
    \item Average detection delay (ADD): $\frac{\sum_{r=1}^R\textrm{max}(d_r - t_s,0)}{\sum_{j=1}^R I(d_j\geq t_s)}$, the average delay among true positive detections only.
\end{itemize}
Figure ~\ref{fig:logbf_trajectories_windowed} shows the collected log-BF sequences with window $w=100$ for the STRS microservice, under the four drift scenarios.  We see the following trends:
\begin{itemize}[noitemsep]
    \item log-BF decreases below 0 for $t< t_s$ when no drift exists.
    \item For $t\in t_s,\dots,t_s+w$ the log-BF increases gradually as the window of observations used to calculate the BF (Section~\ref{ssec:windowed_BF}) gradually includes a greater proportion of drifted vectors $C_t,\:t\geq t_s$.
    \item When the BF sample size is at the full window size and fully non-drifted ($t\in[w,t_s-1]$) or drifted ($t\in[t_s+w-1,t_s+\ell]$), the log-BF is roughly constant.
    \item If contamination ends ($t>t_s+\ell$), the log-BF gradually reverses to the initial level when $t<t_s$.
    \item The log-BF reaches a higher value when the contamination is higher ($p=0.3$ vs 0.1).
\end{itemize}

Figure~\ref{fig:logbf_trajectories_full} shows plots analogous to Figure~\ref{fig:logbf_trajectories_windowed} on repeated experiments with the same settings, except that the full past history is used ($w=\infty$).  We see that

\begin{itemize}[noitemsep]
    \item During the pre-contamination span $t<t_s$, the log-BF continues to decline, reflecting the increased evidence in favor of the prior (since the window size is not limited) rather than flattening.
    \item Similarly, the log-BF continues to rise during the contamination injection period, rather than flattening after $w=100$ observations.
    \item After contamination ends, the decrease in the BF is slower than in the windowed case.
\end{itemize}

\def\smallpanelwidth{0.8}

\begin{figure*}
\centering
\includegraphics[width=\smallpanelwidth\columnwidth]{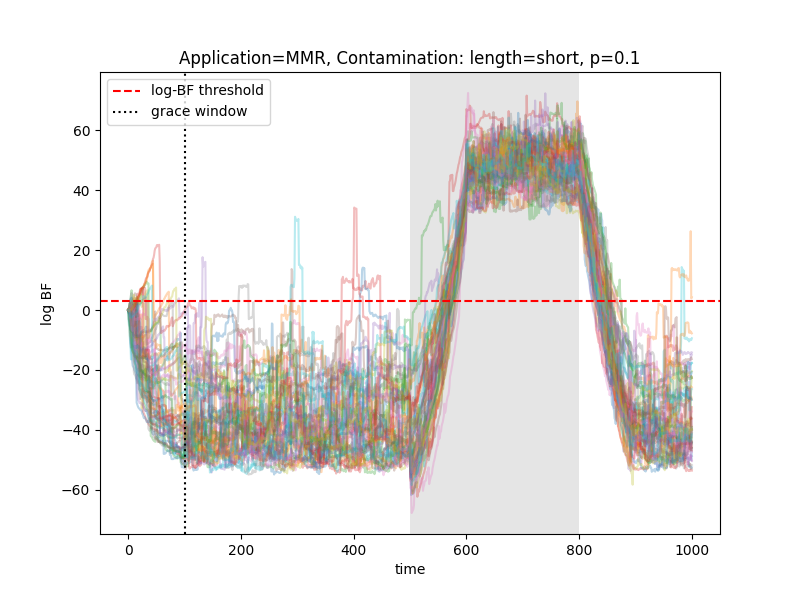}
\includegraphics[width=\smallpanelwidth\columnwidth]{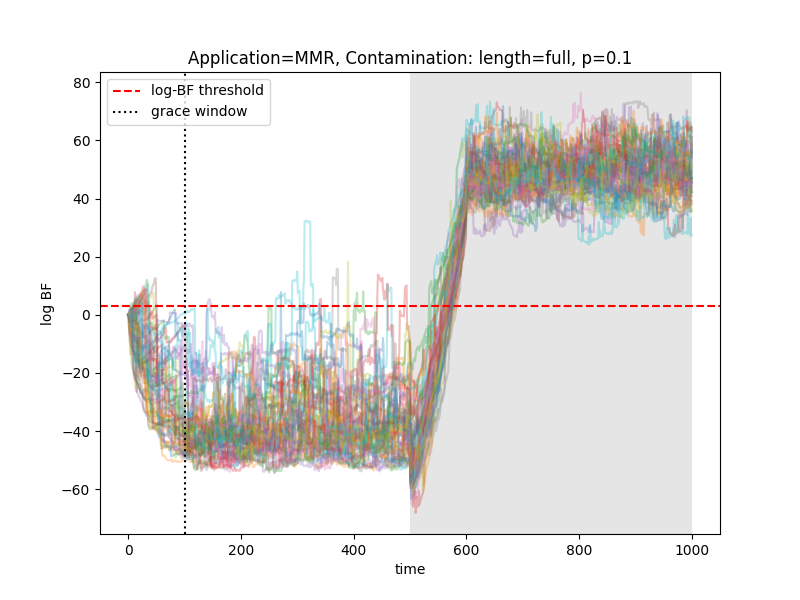}
\includegraphics[width=\smallpanelwidth\columnwidth]{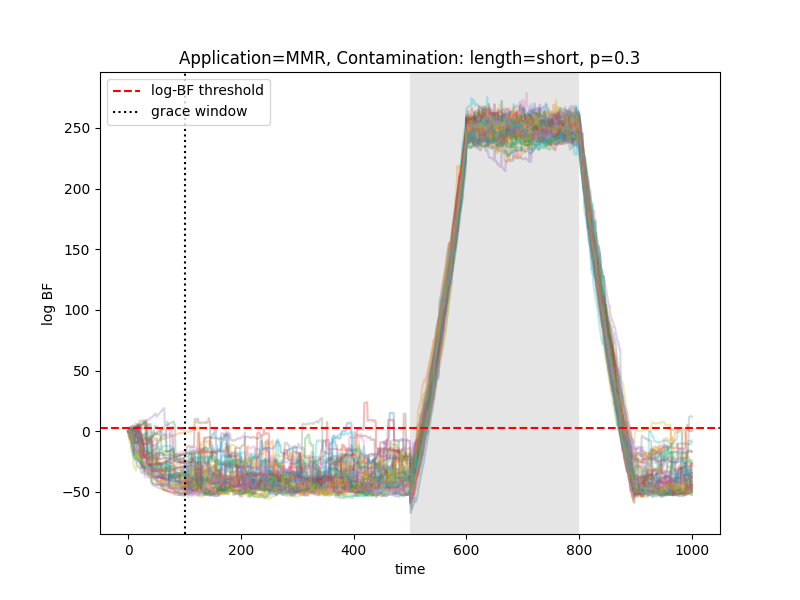}
\includegraphics[width=\smallpanelwidth\columnwidth]{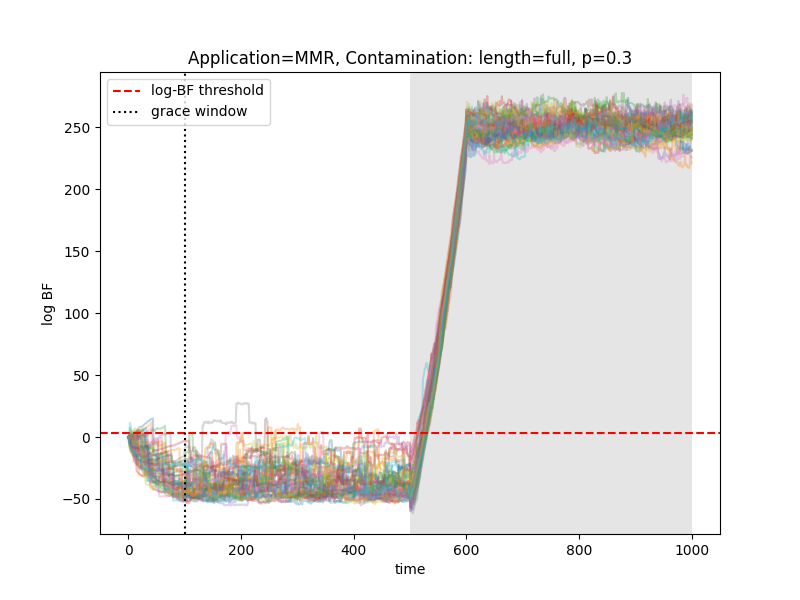}
\caption{Log-BF sequences for $R=50$ simulations of STRS microservice, under four scenarios: Contamination $p=0.1$ (top) and $p=0.3$ (bottom); drift length $\ell=150$ (`short', left) and $\ell=\infty$ (`long', right). The grey background shows the time span during which contamination was injected ($t\colon\: p_t>0$).
\label{fig:logbf_trajectories_windowed}
}
\end{figure*}

\begin{figure*}
\centering
\includegraphics[width=\smallpanelwidth\columnwidth]{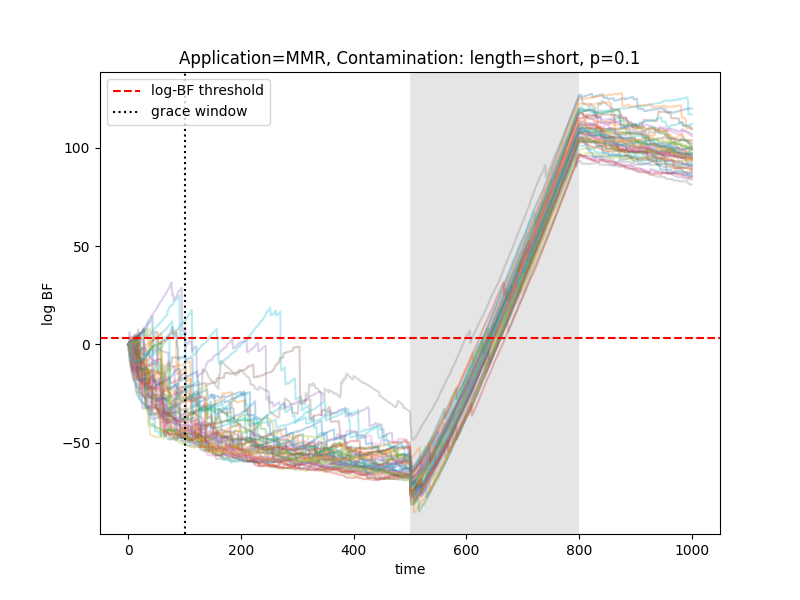}
\includegraphics[width=\smallpanelwidth\columnwidth]{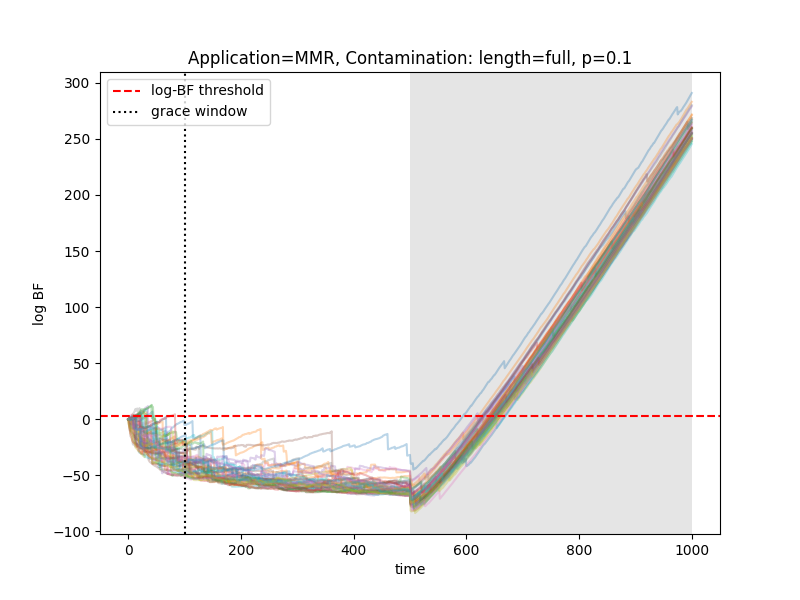}
\includegraphics[width=\smallpanelwidth\columnwidth]{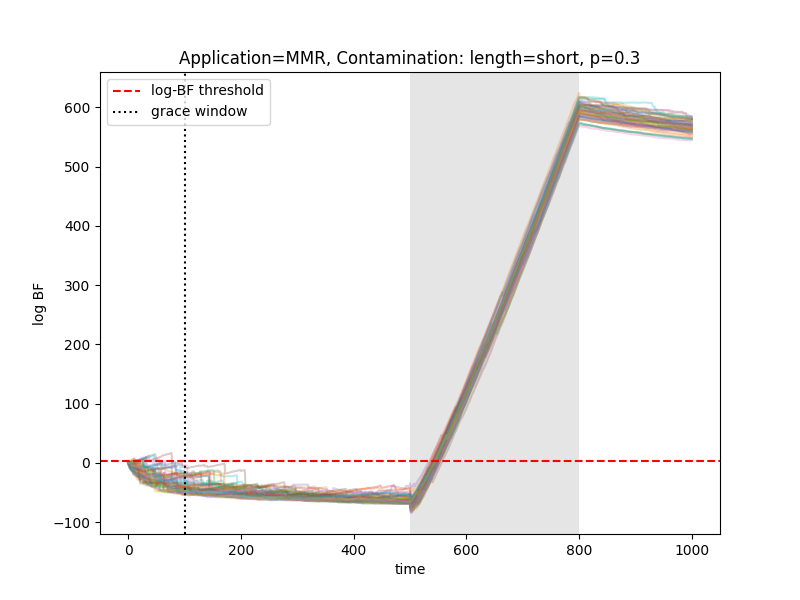}
\includegraphics[width=\smallpanelwidth\columnwidth]{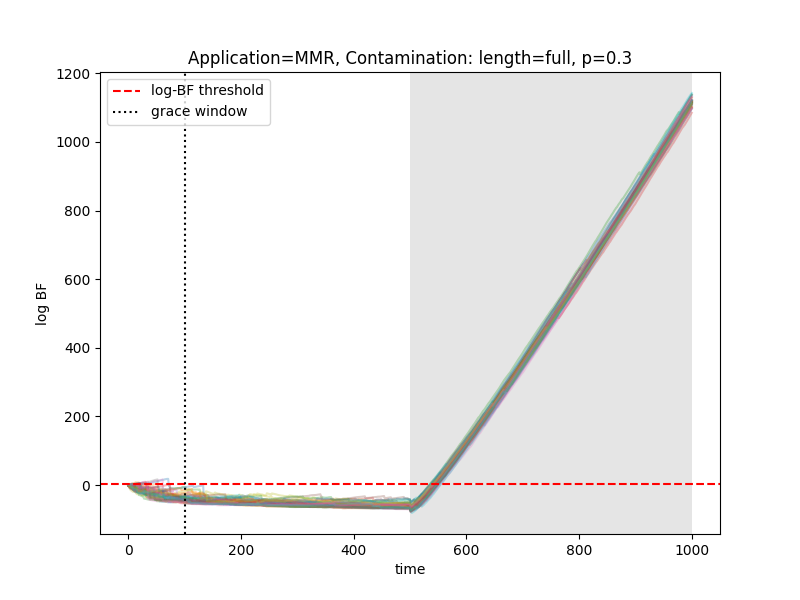}
\caption{Log-BF sequences when no windowing is used ($w=\infty$).
\label{fig:logbf_trajectories_full}
}
\end{figure*}

Table~\ref{tab:simulation_summaries} contains summary statistics of the simulations.  The top of the table has $w=100$ for windowing, and the bottom rows use the full history to calculate the BF.  Using the full history gives lower FPR, but with significantly delayed correct decisions.  Furthermore, the BF will be less responsive to changes.  However, this experiment may be too restrictive by only considering the first detection, which may be a FP.  The results could be improved by raising the threshold or using a human-in-the-loop to inspect detections, which would make a few FPs less worrisome.

\begin{table}\begin{tabular}{|c|c | c c c c |}
\hline
Service & \thead{ Window $w$ } & TPR & FPR & FNR & ADD\\ 
 \hline
 MTR & 100 &\fpeval{round(50/50, 2)} & \fpeval{round(0/50, 2)} & \fpeval{round(0/50, 2)} & \fpeval{round(75.76,0)}\\
 MMR & 100 &\fpeval{round(40/50, 2)} & \fpeval{round(10/50, 2)} & \fpeval{round(0/50, 2)} & \fpeval{round(23.2, 0)} \\
 NLU & 100 &\fpeval{round(7/50, 2)} & \fpeval{round(43/50, 2)} & \fpeval{round(0/50, 2)} & \fpeval{round(35.857142857142854, 0)} \\
 STR & 100 &\fpeval{round(8/50, 2)} & \fpeval{round(42/50, 2)} & \fpeval{round(0.0/50, 2)} & \fpeval{round(54.75, 0)}\\
 STRS & 100 &\fpeval{round(46/50, 2)} & \fpeval{round(4/50, 2)} & \fpeval{round(0/50, 2)} & \fpeval{round(76.1086956521739, 0)}\\
 \hline
 MTR & $\infty$ &\fpeval{round(50/50, 2)} & \fpeval{round(0/50, 2)} & \fpeval{round(0/50, 2)} & \fpeval{round(186.66, 0)}\\
 MMR & $\infty$ &\fpeval{round(50/50, 2)} & \fpeval{round(0/50, 2)} & \fpeval{round(0/50, 2)} & \fpeval{round(44.5, 0)} \\
 NLU & $\infty$ &\fpeval{round(35/50, 2)} & \fpeval{round(15/50, 2)} & \fpeval{round(0/50, 2)} & \fpeval{round(66.85714285714286, 0)}\\
 STR & $\infty$ &\fpeval{round(39/50, 2)} & \fpeval{round(11/50, 2)} & \fpeval{round(0/50, 2)} & \fpeval{round(129.7948717948718, 0)}\\
 STRS & $\infty$ &\fpeval{round(50/50, 2)} & \fpeval{round(0/50, 2)} & \fpeval{round(0/50, 2)} & \fpeval{round(200.32, 0)}\\
 \hline
\end{tabular}
\caption{Summary on drift simulation for a scenario with contamination $p=0.3$ and full contamination length. The top rows show results using the windowed BF, and the bottom is if the full history is used.
\label{tab:simulation_summaries}}
\end{table}

\subsection{Non-simulated data
\label{ssec:nonsimulated}
}
We have an instrumented application, Quote of The Day\footnote{\url{https://gitlab.com/quote-of-the-day/quote-of-the-day}} (QoTD) which demonstrates a real-time infrastructure setup. The QoTD app has 8 microservices and a database service. This application has a suite of testing tools, like load generation and anomaly generation which are helpful in testing the LAD pipeline. The LAD pipeline is trained on the logs generated by these services. The prior multinomial vector of non-drifted logs---the $\prior{\boldsymbol{\alpha}}{}$---is calculated during this training phase. 
 
For inference, we introduce cascading failures, which start by introducing latency in 2 services, cascading to 3 other services, and generating error codes in API responses and service failures. These logs are when we expect the drift detection model to flag specific applications to be drifted.

\begin{figure}
\centering
\includegraphics[width=\columnwidth]{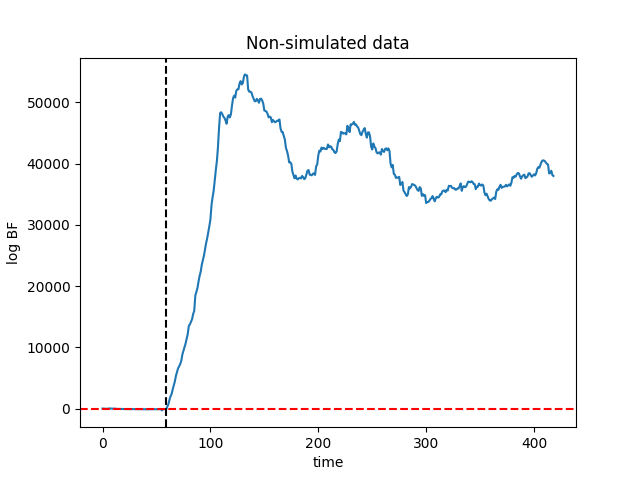}
\caption{Log-BF sequences for QoTD application logs\label{fig:logbf_nonsimulated}
}
\end{figure}

We evaluate the algorithm on a set of 60 training vectors $C_N$ followed by 360 anomalous vectors $C_A$.  Instead of synthetically simulating drift mixtures as in Section~\ref{ssec:cv_drift_simulation}, we simply monitor the concatenated sequence $C=C_N+C_A$, and thus $t_s=61$; the window size is $w=50$.  The resulting log-BF is shown in Figure~\ref{fig:logbf_nonsimulated}.  There are two early false positive detections occurring when there is initial instability in the BF, which can be ignored if we allow a grace window.  Then, the log-BF increases significantly after the anomalous vectors are observed.

\section{Future Work}

In future work, we intend to expand on the following aspects introduced in our detection algorithm.

\begin{itemize}[noitemsep]
    \item Considering more contamination injection scenarios, such as gradual increases in the level, rather than remaining constant (Section~\ref{ssec:cv_drift_simulation}).
    \item Determining the optimal BF window $w$.
    \item Correcting the BF detection threshold, if necessary, to reflect the fact that the moving window forgets past observations.
\end{itemize}

    


\section{Conclusions
\label{sec:conclusions}}

In this work, we present a drift detection technique and its seamless integration into a Log Anomaly Detection pipeline. This integration serves the purpose of recognizing alterations in system activity by examining fluctuations in the occurrence of feature vectors represented as observed log templates. The drift detection procedure employs  Bayesian inference and incorporates a windowing component to enhance its responsiveness to the most recent changes in activity. While we apply our approach within a specific LAD pipeline, the method is versatile enough to be utilized in other settings with problem-specific feature engineering.


\bibliography{main.bib}

\newpage




\end{document}